\documentclass[10pt,twocolumn,letterpaper]{article}

\usepackage[pagenumbers]{cvpr} 
\usepackage{multirow}
\usepackage{units}
\usepackage{float}
\usepackage{placeins}
\usepackage{amssymb}
\usepackage{pifont}
\definecolor{cvprblue}{rgb}{0.21,0.49,0.74}
\usepackage[pagebackref,breaklinks,colorlinks,allcolors=cvprblue]{hyperref}

\definecolor{lowred}{RGB}{238,18,137}
\newcommand{\tabincell}[2]{\begin{tabular}{@{}#1@{}}#2\end{tabular}}
\newcommand{\dplus}[1]{\fontsize{5pt}{0.1em}\selectfont (\textbf{\textcolor{lowred}{#1}})}
\def\paperID{xxx} 
\def\confName{CVPR}
\def\confYear{2025}

\title{A Unified Image-Dense Annotation Generation Model for Underwater Scenes}

\author{
Hongkai Lin
\quad Dingkang Liang
\quad Zhenghao Qi
\quad Xiang Bai\textsuperscript{*}
\\
Huazhong University of Science and Technology\\
{\tt \{hklin,dkliang,xbai\}@hust.edu.cn}
}

\begin{document}
\twocolumn[{%
\maketitle
\begin{center}
\vspace{-20pt}
    \captionsetup{type=figure}
    \includegraphics[width=0.95\textwidth]{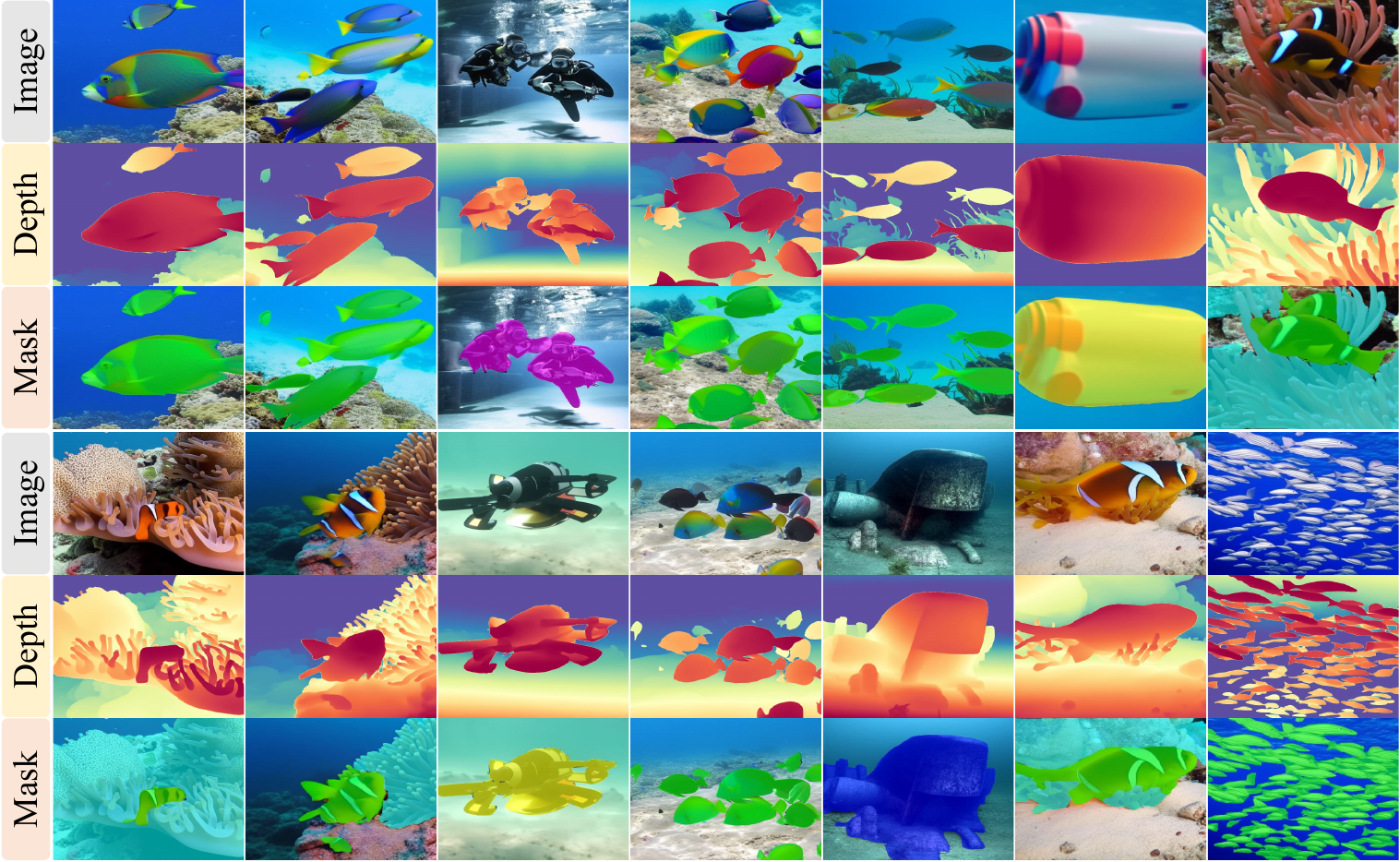}
    \vspace{-5pt}
    \captionof{figure}{\textbf{We present TIDE, a unified underwater image-dense annotation generation model.} 
    Its core lies in the shared layout information and the natural complementarity between multimodal features.
    Our model, derived from the text-to-image model and fine-tuned with underwater data, enables the generation of highly consistent underwater image-dense annotations from solely text conditions.}
    \label{fig:SynTIDE}
\end{center}
}]
{\let\thefootnote\relax\footnotetext{* Corresponding author.}}

\begin{abstract}
Underwater dense prediction, especially depth estimation and semantic segmentation, is crucial for gaining a comprehensive understanding of underwater scenes.
Nevertheless, high-quality and large-scale underwater datasets with dense annotations remain scarce because of the complex environment and the exorbitant data collection costs. 
This paper proposes a unified \textbf{T}ext-to-\textbf{I}mage and \textbf{DE}nse annotation generation method (TIDE) for underwater scenes. 
It relies solely on text as input to simultaneously generate realistic underwater images and multiple highly consistent dense annotations. 
Specifically, we unify the generation of text-to-image and text-to-dense annotations within a single model.
The \textbf{I}mplicit \textbf{L}ayout \textbf{S}haring mechanism (ILS) and cross-modal interaction method called \textbf{T}ime \textbf{A}daptive \textbf{N}ormalization (TAN) are introduced to jointly optimize the consistency between image and dense annotations.
We synthesize a large-scale underwater dataset using TIDE to validate the effectiveness of our method in underwater dense prediction tasks.
The results demonstrate that our method effectively improves the performance of existing underwater dense prediction models and mitigates the scarcity of underwater data with dense annotations.
We hope our method can offer new perspectives on alleviating data scarcity issues in other fields.
The code is available at \url{https://github.com/HongkLin/TIDE}.
\end{abstract}

\section{Introduction}
\label{sec:intro}
Underwater dense prediction, particularly depth estimation and semantic segmentation, is essential for underwater exploration and environmental monitoring. 
However, the complex environment and the prohibitive data collection costs result in a scarcity of underwater data with dense annotations.
Such conditions severely hinder the advancement of dense prediction technologies in underwater scenes.

Fortunately, the recent success of the image generative technique~\cite{ho2020denoising, rombach2022high, zhang2023adding} provides a breakthrough in addressing the scarcity of underwater scene data.
In the field of general object understanding, controllable data synthesis~\cite{wang2024detdiffusion, yang2024freemask, jiang2025minima, nguyen2024dataset} demonstrates its effectiveness in few-shot scenarios.
A straightforward solution is to apply them to underwater scenes directly.
For instance, Atlantis~\cite{zhang2024atlantis}, a pioneering controllable generation method for underwater depth data that takes ControlNet as its core, utilizes terrestrial depth maps as conditions. 
It effectively mitigates the issue of scarce underwater depth data and achieves consistent performance improvements across multiple underwater depth datasets and models.

\begin{figure}[t]
  \centering
   \includegraphics[width=1.0\linewidth]{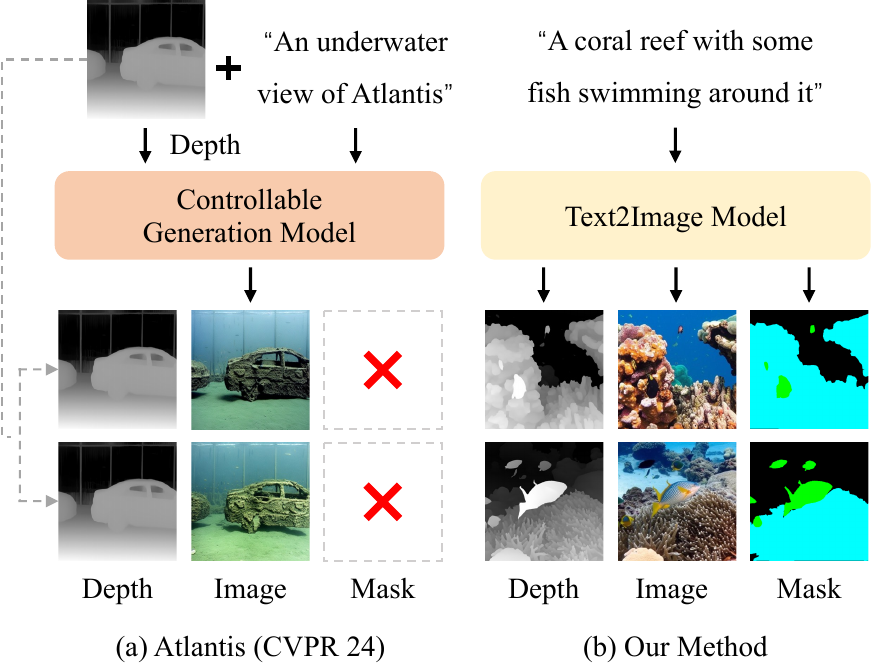}

   \caption{The comparison between Atlantis~\cite{zhang2024atlantis} and our method. Unlike Atlantis, which requires text and depth map conditions, our method only needs text as the input condition to generate image-dense annotations (e.g., depth maps and semantic masks).}
   \label{fig:AtlantisVSTIDE}
\end{figure}

Despite remarkable progress, there are still challenges in Atlantis, as follows:
1) Atlantis, as shown in Fig.~\ref{fig:AtlantisVSTIDE}(a), generates underwater depth data using terrestrial depth maps as conditions due to the lack of underwater depth maps.
It is considered a suboptimal approach since it may not align with natural underwater scenes.
Better recreating authentic underwater environments is equally essential.
2) It generates data with only a single type of dense annotations, which is insufficient for understanding complex underwater scenes.
Thus, a natural question arises: \textit{How can we simultaneously generate highly consistent, one-to-many, and vivid underwater images and dense annotation pairs?}

In this paper, we explore the possibility of simultaneously generating highly consistent, realistic underwater scene images and multiple types of dense annotations using only text conditions.
Our approach, which we refer to as \textbf{TIDE}, is illustrated in Fig.~\ref{fig:AtlantisVSTIDE}(b), presents a unified \textbf{T}ext-to-\textbf{I}mage and \textbf{DE}nse annotation generation method.
TIDE is an end-to-end training and inference model that integrates denoising models in parallel for both text-to-image generation and text-to-dense annotation generation.

To align the images and multiple type dense annotations generated by parallel denoising models, we propose the \textbf{I}mplicit \textbf{L}ayout \textbf{S}haring (ILS) mechanism.
Specifically, the cross-attention map as an implicit layout is the key to controlling the image layout in the text-to-image model~\cite{rombach2022high, chen2023pixartalpha}, inspiring us to share the implicit layout for aligning images and dense annotations.
ILS effortlessly replaces the cross-attention map in the text-to-dense annotation model with that from the text-to-image model, effectively improving the consistency between the image and dense annotations.
Furthermore, considering the intrinsic complementarity between features of different modalities, we introduce a cross-modal interaction method called \textbf{T}ime \textbf{A}daptive \textbf{N}ormalization (TAN),
a normalization layer that modulates the activations using different modal features.
The consistency of the image and dense annotations can further be jointly optimized through cross-modal feature interaction among different dense annotation generation and between image and dense annotation generation.

To verify the effectiveness of our method, we use TIDE to generate a large-scale dataset of underwater images with dense annotations named SynTIDE.
Extensive experiments demonstrate the effectiveness of SynTIDE for underwater dense prediction tasks.
In the underwater depth estimation task, SynTIDE presents consistent improvements in various fine-tuning models.
For example, when adopting representative NewCRFs~\cite{yuan2022neural} as the fine-tuning model, our approach achieves significance gains over previous work, particularly in the $SI_{log}$ and $\delta_1$ metrics, with improvements of $14.73$ and $36\%$ on the D3 and D5 subsets of Sea-thru~\cite{akkaynak2019sea} dataset, respectively.
In underwater semantic segmentation, pre-training with SynTIDE yields consistent improvements across different models.
For instance, when using ViT-Adapter~\cite{chenvision} as the training model, pre-training with the SynTIDE dataset leads to improvements of $2.1\%$ mIoU on the USIS10K~\cite{lian2024diving} dataset.

TIDE demonstrates powerful data generation capabilities for underwater scenes.
Using only easily accessible text prompts, TIDE can generate highly consistent and realistic underwater images and multiple types of dense annotations.
It holds potential as a mainstream data synthesis method for underwater scenes and offers a promising direction for alleviating data scarcity in other fields.
The main contributions of this work are as follows:
\textbf{1)} We propose a novel data synthesis method, TIDE, which uses text as the sole condition to generate images and their corresponding multi-type dense annotations simultaneously.
To our knowledge, TIDE is the first method capable of simultaneously synthesizing both images and multiple dense annotations from text.
\textbf{2)} To align the images and dense annotations, we introduce the Implicit Layout Sharing mechanism. The text-to-image and text-to-dense annotation models share the same layout information, ensuring proper alignment between the image and dense annotations.
Meanwhile, the consistency between image and dense annotations can be further optimized through the cross-modal interaction method called Time Adaptive Normalization.

\section{Relate Work}
\label{sec:relate work}
\subsection{Underwater Dense Prediction} 
Dense prediction tasks in underwater scenes are crucial for comprehensively understanding underwater scenes.
The publication of SUIM~\cite{islam2020semantic} provides a fundamental dataset and benchmark for the exploration of underwater semantic segmentation.
To fill the gap in underwater instance segmentation, WaterMask~\cite{lian2023watermask} publishes the UIIS dataset, and a model is designed to cater to the unique characteristics of underwater images, improving the accuracy of underwater instance segmentation.
Recently, the rise of general foundational segmentation models ~\cite{kirillov2023segment, ravi2024sam} drives further development in the field of underwater segmentation~\cite{lian2024diving, yan2024mas, zhang2024fantastic}.

Due to the lack of underwater depth estimation datasets, most underwater depth estimation methods focus on traditional techniques, unsupervised, or self-supervised approaches.
Traditional methods~\cite{drews2016underwater} mainly rely on statistical priors, such as the dark channel prior~\cite{he2010single}, to estimate underwater depth. 
Gupta et al.~\cite{gupta2019unsupervised} model the relationship between underwater and above-water hazy appearances to depth estimation. 
UW-GAN~\cite{hambarde2021uw} and Atlantis~\cite{zhang2024atlantis} improve the performance of underwater depth estimation by synthesizing training datasets through generative models.

While these methods make notable contributions to underwater dense prediction tasks, the large-scale and high-quality dataset in underwater scenes with only segmentation or depth annotations remains insufficient for achieving comprehensive underwater scene understanding.

\subsection{Controllable Data Synthesis}
Thanks to the success of diffusion models~\cite{ho2020denoising} and the availability of large-scale, high-quality text-image training data, text-to-image models~\cite{nichol2022glide, rombach2022high, ramesh2022hierarchical, chen2023pixartalpha} and controllable image generation models~\cite{zhang2023adding, xue2023freestyle, lv2024place} achieve unprecedented success in image quality, diversity, and consistency.

He et al.~\cite{he2023synthetic} are the first to explore and demonstrate the effectiveness of state-of-the-art text-to-image generation models for image recognition.
This makes it possible to achieve diverse data collection and accurate annotation at a lower cost.
Wu et al. and Nguyen et al.~\cite{wu2023datasetdm, nguyen2024dataset} explore the ability of pre-trained diffusion models to enhance real data in few-shot settings for segmentation tasks.
Diffumask~\cite{wu2023diffumask} ingeniously combines text-to-image models with AffinityNet~\cite{ahn2018learning}, achieving open-vocabulary segmentation data synthesis.
Freemask~\cite{yang2024freemask} demonstrates that synthetic data can further enhance the performance of semantic segmentation models under fully supervised settings by incorporating freestyle, a controllable image generation method using semantic masks as input conditions.
Seggen~\cite{ye2023seggen} designs a multi-stage semantic segmentation data synthesis method, text2mask and mask2image, which achieves high semantic consistency semantic segmentation data only using text as the condition.
Detdiffusion~\cite{wang2024detdiffusion} synthesizes object detection data by incorporating object categories and spatial coordinates into the text.

Unlike the aforementioned single-task data synthesis methods, we propose a novel end-to-end underwater data synthesis approach that simultaneously generates semantic masks and depth maps, relying solely on text conditions.
\begin{figure*}[t]
  \centering
   \includegraphics[width=1.0\linewidth]{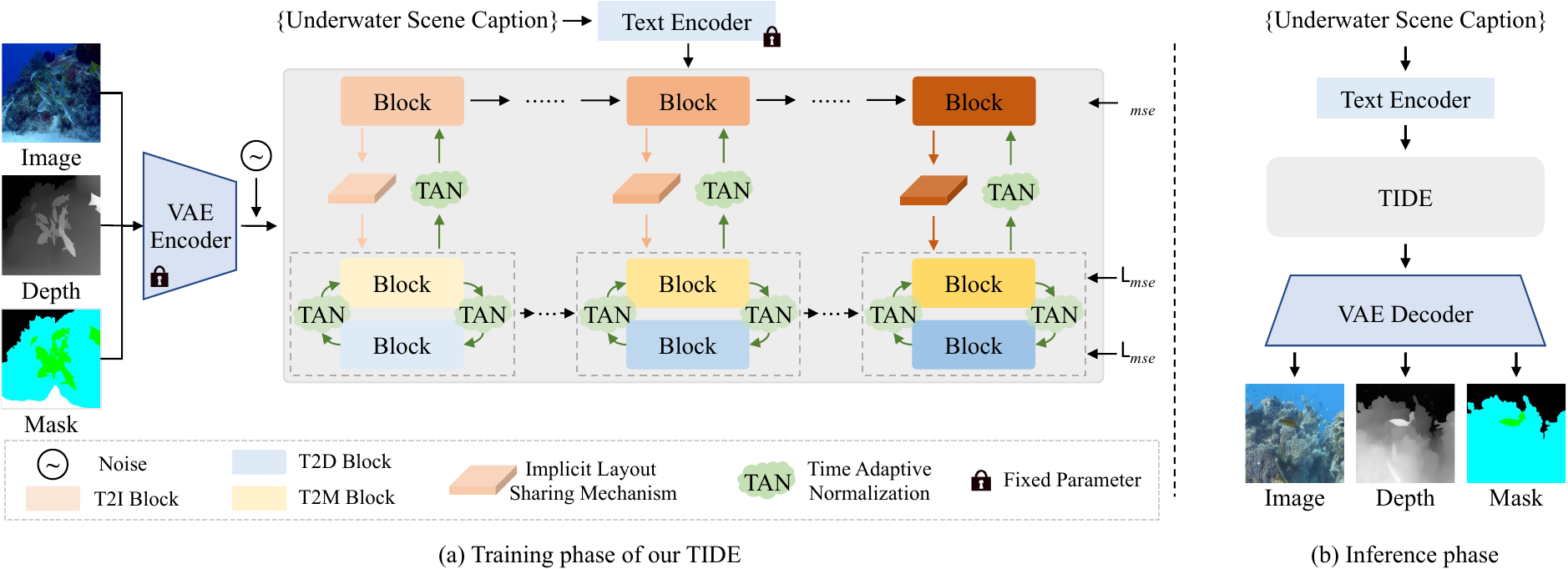}
   \caption{Training and Inference. The denoising model of TIDE mainly consists of three transformers, each dedicated to text-to-image, text-to-depth, and text-to-mask. The proposed Implicit Layout Sharing mechanism (ILS) and Time Adaptive Normalization (TAN) are used to align the generated image, depth map, and semantic mask.}
   \label{fig:training_inference}
\end{figure*}
\section{Preliminaries}
\label{sec:preliminaries}
Diffusion Models (DMs)~\cite{ho2020denoising} emerge as leading text-to-image (T2I) generation models, recognized for their ability to produce realistic images. 
DMs can reconstruct data distribution by learning the reverse process of a diffusion process.
Denoting $z_{t}$ as the random variable at $t$-th timestep, the diffusion process is modeled as a Markov Chain:

\begin{equation}
    \label{eq:markov-chain}
    z_{t} \sim  \mathcal{N}(\sqrt{\alpha_{t}}z_{t-1}, (1-\alpha_{t})\textbf{\textit{I}})\text{,} 
\end{equation} 
\noindent where $\alpha_{t}$ is the fixed coefficient predefined in the noise schedule, and $\textbf{\textit{I}}$ refers to identity matrix. 
A prominent variant, the Latent Diffusion Model (LDM)~\cite{rombach2022high}, innovatively shifts the diffusion process of standard DMs into a latent space. 
This transition notably decreases computational costs while preserving the generative quality and flexibility of the original model. 
The resulting efficiency gain primarily arises from the reduced dimensionality of the latent space, which allows for lower training costs without compromising the model's generative capabilities.

Stable Diffusion, an exemplary implementation of LDM, comprises an AutoEncoder~\cite{van2017neural} and a latent diffusion model.
The AutoEncoder \scalebox{1.5}{$\varepsilon$} is designed to learn a latent space that is perceptually equivalent to the image space. 
Meanwhile, the LDM \scalebox{1.5}{$\epsilon_{\theta}$} is parameterized as a denoising model with cross-attention and trained on a large-scale dataset of text-image pairs via:
\begin{equation}
    \label{eq:denoising}
    \mathcal{L}_{LDM}:=\mathbb{E}_{\varepsilon(x),y,\epsilon\sim N(0,1),t}[\|\epsilon - \epsilon_{\theta}(z_{t}, t, \tau_{\theta}(y)) \|_{2}^{2}]\text{,} 
\end{equation} 
\noindent where $\epsilon$ is the target noise.
$\tau_{\theta}$ and $y$ are the pre-trained text encoder (e.g., CLIP~\cite{radford2021learning}, T5~\cite{raffel2020exploring}) and text prompts, respectively.
This equation represents the mean-squared error (MSE) between the target noise $\epsilon$ and the noise predicted by the model, encapsulating the core learning mechanism of the latent diffusion model.

\section{Our Method}
\label{sec:methods}
An overview of our method, a unified text-to-image and dense annotation generation model (TIDE), is shown in Fig.~\ref{fig:training_inference}.
TIDE is built upon a pre-trained transformer~\cite{chen2023pixartalpha} for text-to-image generation, along with two fine-tuned mini-transformers (details provided in Sec.~\ref{sec:imple}) dedicated to text-to-depth and text-to-mask generation.
Simply parallelizing multiple text-to-image processes does not ensure consistency between the images and dense annotations.
To enable consistency between them, we propose Implicit Layout Sharing (ILS) and the cross-modal interaction method named Time Adaptive Normalization (TAN).
After training, TIDE simultaneously generates images and multiple dense annotations with high consistency using only text as input.
\subsection{Data Preparation}
\label{subsec:data_preparation}
We aim to generate realistic underwater images, corresponding highly consistent depth maps, and semantic masks.
However, existing high-quality, dense annotation data primarily consists of mask annotations. 
Therefore, we construct training data around these datasets with semantic masks, as shown in Tab.~\ref{tab:datasource}. 
On this basis, we obtain the corresponding depth map and caption for each image using existing foundation models.
Specifically, for each underwater image, the corresponding depth map is obtained by pre-trained Depth Anything~\cite{yang2024depthv2}. 
Meanwhile, the caption of each image is obtained from the pre-trained BLIP2~\cite{li2023blip}.
We construct approximately 14K quadruples \{Image, Depth, Mask, Caption\} for TIDE training. 
\begin{table}[h]
  \centering
  \footnotesize
  \setlength{\tabcolsep}{3.7mm}
    \caption{Segmentation Datasets and Data Splits. $^{\star}$ denotes the training set of TIDE, while the others are used for evaluation.}
  \begin{tabular}{@{}lcccc@{}}
    \toprule
    Datasets & Seg Task & Train & Val & Test \\
    \midrule
    SUIM~\cite{islam2020semantic} & Semantic & $1,488^{\star}$ & $110^{\star}$& /   \\
    UIIS~\cite{lian2023watermask} & Instance & $3,937^{\star}$ & $691$  & /            \\
    USIS10K~\cite{lian2024diving} & Instance & $7,442^{\star}$ & $1,594$ & $1,596^{\star}$\\
    \bottomrule
  \end{tabular}
  \label{tab:datasource}
\end{table}
\subsection{Implicit Layout Sharing Mechanism} 
\label{subsec: shared}
In advanced text-to-image models~\cite{rombach2022high, chen2023pixartalpha}, the cross-attention map plays a crucial role in controlling the image layout.
Existing methods~\cite{xue2023freestyle, lv2024place} demonstrate that adjusting the cross-attention map during the text-to-image process can effectively control the layout of the generated image.
Therefore, the cross-attention map can be considered as the implicit layout information.
Intuitively, sharing the implicit layout between text-to-image and text-to-dense annotations may establish a strong correlation between the generated image and dense annotations.
To this end, we propose an Implicit Layout Sharing mechanism to align the generated image and dense annotations.
Specifically, cross-attention, as a crucial process for generating implicit layouts in text-to-image/mask/depth model, can first be formulated as:
\begin{equation} 
    \label{eq:depth_oca}
    \begin{aligned} 
        \mathtt{Attn}_{i}(\boldsymbol{Q}_{i}, \boldsymbol{K}_{i}, \boldsymbol{V}_{i}) &= \mathtt{softmax}(
        \nicefrac{
        \boldsymbol{Q}_{i} \boldsymbol{K}_{i}^{\top}
        }{\sqrt{c}}
        )
        \boldsymbol{V}_{i}\text{,}\\
        \mathtt{Attn}_{d}(\boldsymbol{Q}_{d}, \boldsymbol{K}_{d}, \boldsymbol{V}_{d}) &= \mathtt{softmax}(
        \nicefrac{
        \boldsymbol{Q}_{d} \boldsymbol{K}_{d}^{\top}
        }{\sqrt{c}}
        )
        \boldsymbol{V}_{d}\text{,}\\
        \mathtt{Attn}_{m}(\boldsymbol{Q}_{m}, \boldsymbol{K}_{m}, \boldsymbol{V}_{m}) &= \mathtt{softmax}(
        \nicefrac{
        \boldsymbol{Q}_{m} \boldsymbol{K}_{m}^{\top}
        }{\sqrt{c}}
        )
        \boldsymbol{V}_{m}\text{,}
    \end{aligned} 
\end{equation}
\noindent where $c$ refers to the feature channel. \(\boldsymbol{Q}_{i}\)/\(\boldsymbol{Q}_{d}\)/\(\boldsymbol{Q}_{m}\), \(\boldsymbol{K}_{i}\)/\(\boldsymbol{K}_{d}\)/\(\boldsymbol{K}_{m}\), and \(\boldsymbol{V}_{i}\)/\(\boldsymbol{V}_{d}\)/\(\boldsymbol{V}_{m}\) represent the query, key, and value within the text-to-image/depth/mask cross-attention module, respectively.
Since text-to-image models are pre-trained on high-quality and large-scale image-caption datasets, they exhibit strong controllability and generalization.
Therefore, sharing the implicit layouts from the text-to-image model is the optimal choice to ensure the quality of the generated data.
As shown in Fig.~\ref{fig:training_inference}(a), the implicit layouts from the block in the text-to-image model are shared with the cross-attention in the block of text-to-dense annotation models. 
The implicit layouts refer to:
\begin{equation}
    \label{eq:image_cam}
    \begin{aligned}
        \boldsymbol{M}_{i} &= \mathtt{softmax}(
        \nicefrac{
        \boldsymbol{Q}_{i} \boldsymbol{K}_{i}^{\top}
        }
        {\sqrt{c}}
        )\text{.}
    \end{aligned}
\end{equation} 
By sharing the implicit layouts from the text-to-image model, the cross-attention of text-to-depth ($\mathtt{Attn}_{d}$) and text-to-mask ($\mathtt{Attn}_{m}$) can be simplified as follows:
\begin{equation} 
    \label{eq:depth_ca}
    \begin{aligned} 
        \mathtt{Attn}_{d}(\boldsymbol{Q}_{d}, \boldsymbol{K}_{d}, \boldsymbol{V}_{d}) &= \boldsymbol{M}_{i} \times \boldsymbol{V}_{d}\text{,}\\
        \mathtt{Attn}_{m}(\boldsymbol{Q}_{m}, \boldsymbol{K}_{m}, \boldsymbol{V}_{m}) &= \boldsymbol{M}_{i} \times \boldsymbol{V}_{m}\text{,}
    \end{aligned} 
\end{equation}

\noindent where $\times$ refers to matrix multiplication. 
Implicit Layout Sharing is an elegant and efficient method that unifies image and dense annotation generation, improving consistency between them.
It also reduces the overall generation cost, as there is no need to compute separate cross-attention maps for the text-to-dense annotation models.

\subsection{Time Adaptive Normalization} 
\label{subsec: tan}
\begin{figure}[t]
  \centering
   \includegraphics[width=0.9\linewidth]{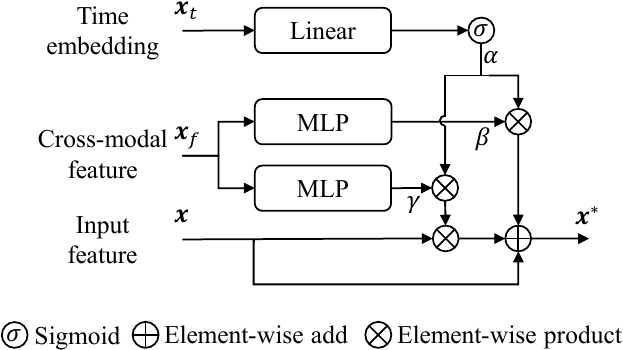}
   \caption{In TAN, the cross-modal features are first mapped to the modulation parameters $\gamma$ and $\beta$. Then, a time-adaptive confidence $\alpha$ is introduced to control the degree of normalization.}
   \label{fig:tan}
\end{figure}

Considering the complementary nature of different modality features, we propose a cross-modal feature interaction method called Time Adaptive Normalization (TAN), as shown in Fig.~\ref{fig:tan}.

Specifically, TAN is utilized to adjust the image layout by leveraging the cross-modal features $\boldsymbol{x}_f$ from different branches. 
The cross-modal features are mapped to two normalization parameters, $\gamma$ and $\beta$, by MLPs, which are used to control the variation in the image layout.
In this context, the features from text-to-depth and text-to-mask serve as cross-modal input features for each other. 
For instance, in the TAN corresponding to the $i$-th text-to-depth block, the outputs from both the $i$-th text-to-depth block and the $i$-th text-to-mask block serve as the input feature $\boldsymbol{x}$ and cross-modal input feature $\boldsymbol{x}_f$, respectively. 
A slight difference is that for text-to-image, the features from both text-to-depth and text-to-mask serve as the cross-modal features.
In the TAN cross-modal interaction process of text-to-image, two sets of $\gamma$ and $\beta$ are obtained, provided by the different modalities features from text-to-depth and text-to-mask. 
These two sets of parameters are averaged to the $\bar{\gamma}$ and $\bar{\beta}$.
Then, time embeddings $\boldsymbol{x}_t$ is introduced to adaptively control the influence of the cross-modal features.
The normalization can be formalized as follows:
\begin{equation} 
    \label{eq:tan}
    \begin{aligned}
        \boldsymbol{x}^{'} = \alpha \cdot \gamma \boldsymbol{x} + \alpha \cdot \beta\text{,}~
        \boldsymbol{x}^{*} = \boldsymbol{x}^{'} + \boldsymbol{x}\text{,}
    \end{aligned}
\end{equation}
\noindent where $\boldsymbol{x}$, $\boldsymbol{x}^{'}$, and $\boldsymbol{x}^{*}$ are the input feature, normalized feature, and output feature, respectively.
$\alpha$ is the time adaptive coefficients obtained from $\boldsymbol{x}_{t}$ through linear transformation and sigmoid. 
The TAN will be applied not only from text-to-dense annotations to text-to-image but also between text-to-depth and text-to-mask to improve the consistency among dense annotations.
Implicit Layout Sharing and Time Adaptive Normalization are two complementary methods that construct a joint interaction process, optimizing the consistency between the generated image and dense annotations during training.

\subsection{Learning Objective} During training, the learnable parameters include only the proposed TAN module and the LoRA~\cite{hu2022lora} used to fine-tune the pre-trained transformer.
The overall loss $\mathcal{L}$ is composed equally of the denoising losses from the three branches: text-to-image, text-to-depth, and text-to-mask:

\begin{equation} 
    \label{eq:loss}
    \mathcal{L} = \mathcal{L}_{mse}^{I} + \mathcal{L}_{mse}^{D} + \mathcal{L}_{mse}^{M}\text{.}
\end{equation}

\subsection{Data Synthesis}
\label{subsec:data synthesis}
Thanks to the proposed ILS and TAN, TIDE can generate realistic and highly consistent underwater images and dense annotations after training, using only text conditions, as shown in Fig.~\ref{fig:training_inference}(b). 

We filter out redundant parts from the 14K captions obtained in Sec.~\ref{subsec:data_preparation}, resulting in approximately 5K non-redundant captions as text conditions.
For each caption, we generate ten samples to construct a large-scale synthetic dataset named SynTIDE. 
Some representative examples are shown in Fig.~\ref{fig:SynTIDE}.
The SynTIDE dataset is utilized to validate the effectiveness of our method in the dense prediction task for underwater scenes. 
\begin{table*}[t]
\renewcommand{\arraystretch}{1.16}
    \centering
    \caption{Quantitative comparisons on real underwater depth estimation datasets.}
    \label{tab:atlantis_vs_tide_seathru}
    \footnotesize
    \setlength{\tabcolsep}{0.8mm}
    \begin{tabular}{@{}cccccccccccc@{}}
    \bottomrule
    Method &Fine-tuning Dataset  & Reference &$SI_{log}\downarrow$ &$A.Rel\downarrow$	&$log_{10}\downarrow$ &$RMSE\downarrow$ &$S.Rel$ ↓&$RMSE_{log}\downarrow$ &$\delta_{1}\uparrow$ &$\delta_{2}\uparrow$ &$\delta_{3}\uparrow$ \\
    \midrule
    \multicolumn{12}{c}{Quantitative comparisons on the D3 and D5 subsets of Sea-thru~\cite{akkaynak2019sea} dataset.} \\
    \midrule
    \multirow{2}{*}{\tabincell{c}{AdaBins~\cite{bhat2021adabins}}} &Atlantis~\cite{zhang2024atlantis} & CVPR 24  &38.24 &1.33 &0.12 &1.41 &\textbf{12.89} &0.39 &0.50 &0.81 &0.92\\
    &SynTIDE (Ours) &-&\textbf{26.92}\dplus{-11.32} &\textbf{1.31}\dplus{-0.02} &\textbf{0.08}\dplus{-0.04} &\textbf{1.12}\dplus{-0.29} &15.74\dplus{+2.85} &\textbf{0.27}\dplus{-0.12} &\textbf{0.71} \dplus{+0.21}&\textbf{0.95}\dplus{+0.14} &\textbf{0.99}\dplus{+0.07} \\
    \midrule
    \multirow{2}{*}{\tabincell{c}{NewCRFs~\cite{yuan2022neural}}} &Atlantis~\cite{zhang2024atlantis} & CVPR 24  &37.10 &1.68 &0.12 &1.44 &\textbf{14.76} &0.38	&0.48 &0.84	&0.95 \\
    &SynTIDE (Ours) &-& \textbf{22.37}\dplus{-14.73} &\textbf{1.50}\dplus{-0.18} &\textbf{0.06}\dplus{-0.06} &\textbf{1.24}\dplus{-0.20} &22.50\dplus{+7.74} &\textbf{0.23}\dplus{-0.15} &\textbf{0.84}\dplus{+0.36} &\textbf{0.97}\dplus{+0.13} &\textbf{0.99}\dplus{+0.04}    \\
    \midrule
    \multirow{2}{*}{\tabincell{c}{PixelFormer~\cite{agarwal2023attention}}} &Atlantis~\cite{zhang2024atlantis} & CVPR 24  &23.70 &\textbf{1.34} &0.06 &1.17 &\textbf{17.29} &0.24 &0.81 &0.97 &\textbf{0.99}\\
    &SynTIDE (Ours) & - &\textbf{21.39}\dplus{-2.31} &1.46\dplus{+0.12} &\textbf{0.05}\dplus{-0.01} &\textbf{1.15}\dplus{-0.02} &21.79\dplus{+4.50} &\textbf{0.22}\dplus{-0.02} &\textbf{0.88}\dplus{+0.07} &\textbf{0.98}\dplus{+0.01} &\textbf{0.99}\dplus{+0.00} \\
    \midrule
    \multirow{2}{*}{\tabincell{c}{MIM~\cite{xie2023revealing}}} &Atlantis~\cite{zhang2024atlantis} & CVPR 24 &37.01 &1.37 &0.11 &1.51 &\textbf{14.42} &0.38 &0.56 &0.84 &0.94 \\
    &SynTIDE (Ours) & - &\textbf{22.49}\dplus{-14.52} &\textbf{1.27}\dplus{-0.10} &\textbf{0.06}\dplus{-0.05} &\textbf{1.01}\dplus{-0.50} &16.46\dplus{+2.04} &\textbf{0.23}\dplus{-0.15} &\textbf{0.85}\dplus{+0.29} &\textbf{0.97}\dplus{+0.13} &\textbf{0.99}\dplus{+0.05} \\
    \midrule
    \multicolumn{12}{c}{Quantitative comparisons on the SQUID~\cite{berman2020underwater} dataset.} \\
    \midrule
    \multirow{2}{*}{\tabincell{c}{AdaBins~\cite{bhat2021adabins}}} &Atlantis~\cite{zhang2024atlantis} & CVPR 24  &29.56 &0.28 &0.11 &\textbf{2.24} &\textbf{0.69} &0.31 &0.56 &0.86 &0.94\\
    &SynTIDE (Ours) &-&\textbf{25.63}\dplus{-3.93} &\textbf{0.23}\dplus{-0.05} &\textbf{0.09}\dplus{-0.02} &2.69\dplus{+0.45} &0.92\dplus{+0.23} &\textbf{0.27}\dplus{-0.04} &\textbf{0.67}\dplus{+0.11} &\textbf{0.90}\dplus{+0.04} &\textbf{0.97}\dplus{+0.03} \\
    \midrule
    \multirow{2}{*}{\tabincell{c}{NewCRFs~\cite{yuan2022neural}}} &Atlantis~\cite{zhang2024atlantis} & CVPR 24  &\textbf{25.19} &\textbf{0.23} &\textbf{0.09} &\textbf{2.56} &\textbf{0.83} &\textbf{0.26} &0.68 &0.90 &0.96 \\
    &SynTIDE (Ours) &-& 25.55\dplus{+0.36} &\textbf{0.23}\dplus{-0.00} &\textbf{0.09}\dplus{+0.00} &3.02\dplus{+0.46} &1.07\dplus{+0.24} &0.27\dplus{+0.01} &\textbf{0.68}\dplus{+0.00} &\textbf{0.91} \dplus{+0.01}&\textbf{0.97}\dplus{+0.01}\\
    \midrule
    \multirow{2}{*}{\tabincell{c}{PixelFormer~\cite{agarwal2023attention}}} &Atlantis~\cite{zhang2024atlantis} & CVPR 24  &21.34 &0.18 &0.07 &1.86 &0.43 &0.22 &0.76 &0.94 &0.98 \\
    &SynTIDE (Ours) & &\textbf{19.08}\dplus{-2.26} &\textbf{0.16}\dplus{-0.02} &\textbf{0.07}\dplus{-0.00} &\textbf{1.75}\dplus{-0.11} &\textbf{0.36}\dplus{-0.07} &\textbf{0.19}\dplus{-0.03} &\textbf{0.79}\dplus{+0.03} &\textbf{0.97}\dplus{+0.03} &\textbf{0.99}\dplus{+0.01} \\
    \midrule
    \multirow{2}{*}{\tabincell{c}{MIM~\cite{xie2023revealing}}} &Atlantis~\cite{zhang2024atlantis} & CVPR 24 &27.45 &0.26 &0.10 &\textbf{2.14} &\textbf{0.68} &\textbf{0.28} &0.61 &0.88 &0.95 \\
    &SynTIDE (Ours) &-&\textbf{26.98}\dplus{-0.47} &\textbf{0.25}\dplus{-0.01} &\textbf{0.09}\dplus{-0.01} &3.04\dplus{+0.90} &1.11\dplus{+0.43} &\textbf{0.28}\dplus{-0.00} &\textbf{0.65}\dplus{+0.04} &\textbf{0.89}\dplus{+0.01} &\textbf{0.96}\dplus{+0.01} \\
    \bottomrule
  \end{tabular}
\end{table*}
\subsection{Analysis}
\noindent \textbf{Insights of framework design.}
In the text-to-image model, the cross-attention map contains the layout information of the image.
Thus, the cross-attention map can be viewed as an implicit layout.
If two text-to-image models share the same implicit layout and undergo proper fine-tuning, the generated images are likely to exhibit strong layout similarity.
Therefore, we share the same implicit layout across multiple text-to-image models.
Meanwhile, we use LoRA to fine-tune the multiple text-to-image models~\cite{chen2023pixartalpha}.

\noindent \textbf{Zero-shot generation ability.}
Thanks to our training strategy, which fine-tunes the pre-trained text-to-image model using only LoRA, the generalization ability of the text-to-image model is retained to some extent.
This enables TIDE to generate underwater images during inference that are not seen during training.
Furthermore, due to the proposed Implicit Layout Sharing and Time Adaptive Normalization mechanisms, the generated depth maps align well with these images. 
Therefore, TIDE has the ability to generate zero-shot underwater image-depth map pairs.
\section{Experiments}
\label{sec:experiments}
\subsection{Dataset and Evaluation Metrics}

\noindent \textbf{Underwater Depth Estimation.}
We follow the work~\cite{zhang2024atlantis}, the D3 and D5 subsets of Sea-thru~\cite{akkaynak2019sea}, and the SQUID dataset~\cite{berman2020underwater} used to evaluate the depth estimation capability in underwater scenes.
These datasets include underwater images with depth maps obtained via the Structure-from-Motion (SfM) algorithm.

The quantitative evaluation metrics include root mean square error ($RMSE$) and its logarithmic variant ($RMSE_{log}$), absolute error in log-scale ($log_{10}$), absolute relative error ($A.Rel$), squared relative error ($S.Rel$), the percentage of inlier pixels ($\delta_{i}$) with thresholds of $1.25^{i}$, and scale-invariant error in log-scale ($SI_{log}$): $100\sqrt{Var(\epsilon_{log})}$.\\[3pt]
\noindent \textbf{Underwater Semantic Segmentation.}
The UIIS~\cite{lian2023watermask} and USIS10K~\cite{lian2024diving} datasets are chosen to validate the effectiveness of our method in underwater semantic segmentation tasks. 
Instance masks belonging to the same semantic category are merged to construct semantic segmentation annotations for the UIIS and USIS10K datasets.

We calculate the mean Intersection over Union (mIoU) for six categories (i.e., Fish, Reefs, Aquatic Plants, Wrecks, Human Divers, and Robots) to evaluate the accuracy of the segmentation results.

\subsection{Implementation Details}
\label{sec:imple}
The training process consists of two parts: pre-training the mini-transformer and training TIDE.
In the first stage, the mini-transformer is initialized with the first ten layers of the PixArt-$\alpha$~\cite{chen2023pixartalpha} pre-trained transformer.
Then, the mini-transformer is trained for 60K iterations on the text-to-image task with all parameters. 
The training data consists of 14K underwater image-caption pairs from Sec.~\ref{subsec:data_preparation}.
In the second stage, the PixArt-$\alpha$ pre-trained transformer and the mini-transformer are used as initial weights for the text-to-image and text-to-dense annotation models, respectively.
Meanwhile, they are fine-tuned using LoRA~\cite{hu2022lora} for 200K iterations with a batch size of 4.
The LoRA ranks of the text-to-image/depth/mask branches are 32, 64, and 64, respectively.
All experiments are conducted on a server with four NVIDIA 4090 24G GPUs.
\begin{table}[t]
    \renewcommand{\arraystretch}{1.43}
    \centering
    \caption{Quantitative results of  underwater semantic segmentation.}
    \label{tab:seg}
    \footnotesize
    \setlength{\tabcolsep}{1.3mm}
    \begin{tabular}{cccccc}
    \toprule
    \multirow{2.3}{*}{Method} & \multirow{2.3}{*}{Backbone}  & \multicolumn{2}{c}{Training Data} & \multicolumn{2}{c}{mIoU} \\
    \cmidrule(lr){3-4} \cmidrule(lr){5-6}
    & & Real & SynTIDE  & UIIS & USIS10K \\ 
    \midrule
    \multirow{3}{*}{\tabincell{c}{Segformer~\cite{xie2021segformer}\\(NeurIPS 21)}}   &\multirow{3}{*}{MiT-B4} &\checkmark & &70.2 &74.6 \\
    & & &\checkmark  &\textbf{76.5} &72.8\\
    & &\checkmark &\checkmark & 75.4\dplus{+5.2}&\textbf{76.1}\dplus{+1.5}\\
    \midrule
    \multirow{3}{*}{\tabincell{c}{Mask2former~\cite{cheng2022masked}\\(CVPR 22)}} &\multirow{3}{*}{Swin-B} & \checkmark & &72.7 & 76.1 \\
    & & &\checkmark&74.2 &72.9 \\
    & &\checkmark &\checkmark &\textbf{74.3}\dplus{+1.6} &\textbf{77.1}\dplus{+1.0} \\
    \midrule
    \multirow{3}{*}{\tabincell{c}{ViT-Adapter~\cite{chenvision}\\(ICLR 23)}} &\multirow{3}{*}{\tabincell{c}{ViT-\\Adapter-B}} & \checkmark & &73.5 &74.6  \\
    & & &\checkmark&\textbf{75.7} &72.6 \\ 
    & &\checkmark &\checkmark &75.1\dplus{+1.6} &\textbf{76.7}\dplus{+2.1}   \\
    \bottomrule
  \end{tabular}
\end{table}
\subsection{Main results}
\noindent \textbf{Underwater Depth Estimation.}
We train four representative depth estimation models, Adasbin~\cite{bhat2021adabins}, NewCRFs~\cite{van2017neural}, PixelFormer~\cite{agarwal2023attention}, and MIM~\cite{xie2023revealing}, to present quantitative results, as shown in Tab.~\ref{tab:atlantis_vs_tide_seathru}.
Compared to previous underwater data synthesis work Atlantis~\cite{zhang2024atlantis}, depth estimation models trained on our SynTIDE dataset show consistent improvements across most quantitative metrics on two evaluated datasets.
Especially on MIM~\cite{xie2023revealing}, a powerful pre-trained model, our method reduces the $SI_{log}$ metric from 37.01 $\rightarrow$ 22.49 (-14.52) and improves $\delta_{1}$ from 0.56 $\rightarrow$ 0.85 (+0.29) on the D3 and D5 subsets of the Sea-thru dataset.
Meanwhile, on PixelFormer~\cite{agarwal2023attention}, a depth estimation model with outstanding generalization that also performs best for Atlantis, our method achieves better performance across nearly all quantitative metrics on both evaluated underwater depth estimation datasets.

These results demonstrate that our method achieves highly competitive consistency compared to Atlantis, which uses stronger dense conditions.
Furthermore, the data generated by TIDE is closer to natural underwater scenes and shows rich species diversity.
Most importantly, TIDE unifies the generation of images and multiple highly consistent dense annotations, capabilities that Atlantis lacks.\\[10pt]
\noindent \textbf{Underwater Semantic Segmentation.}
In the underwater semantic segmentation task, we validate the effectiveness of our method by pre-training with the SynTIDE dataset in three representative semantic segmentation models, Segformer~\cite{xie2021segformer}, Mask2former~\cite{cheng2022masked}, and ViT-Adapter~\cite{chenvision}.
Following the work~\cite{yang2024freemask}, we filter the noise in the generated annotations with 1.5 tolerance.

Pre-training on high-quality synthetic datasets is widely recognized as a way to gain strong prior knowledge. 
On the UIIS dataset, models trained on the SynTIDE dataset consistently achieve superior results compared to real data.
On the other larger USIS10K dataset, by further fine-tuning the model on the UIIS10K train set, we achieve notable improvements. 
Especially on ViT-Adapter, we enhance the performance of the model from 74.6\% $\rightarrow$ 76.7\%.

These results show that models pre-trained on the SynTIDE dataset exhibit strong prior knowledge in the underwater semantic segmentation task. Additionally, these results demonstrate that the unified image and dense annotation generation model proposed in this paper can generate highly consistent image-dense annotation pairs, making it suitable for various underwater dense prediction tasks.
\begin{table}[t]
    \renewcommand{\arraystretch}{1.2}
    \centering
    \caption{Ablation on the impact of each TIDE component.}
    \label{tab:abl_components}
    \footnotesize
    \setlength{\tabcolsep}{3.5mm}
    \begin{tabular}{cccccc}
    \toprule
    ILS & TAN & $SI_{log}\downarrow$ &$A.Rel\downarrow$	 &$\delta_{1}\uparrow$ & mIoU\\
    \midrule
    \checkmark & &24.46 &1.23 &0.76 &36.8 \\
     &\checkmark &24.59 &1.40 &0.78 &36.2  \\
    \checkmark & \checkmark &23.71 &1.37 &0.79 &42.1 \\
    \bottomrule
  \end{tabular}
\end{table}
\begin{table}[t]
    \renewcommand{\arraystretch}{1.2}
    \centering
    \vspace{5pt}
    \caption{Ablation on the impact of the component positions. ``\{Start, End\}" indicates the starting and ending positions where the operations are applied, with a step size of 3.}
    \label{tab:abl_position}
    \footnotesize
    \setlength{\tabcolsep}{3.7mm}
    \begin{tabular}{ccccc}
    \toprule
    \{Start, End\} & $SI_{log}\downarrow$ &$A.Rel\downarrow$	&$\delta_{1}\uparrow$  & mIoU\\
    \midrule
    \{0,12\}  &22.06 &1.46 &0.86 &34.7\\
    \{15,27\} &44.86 &1.09 &0.43 &8.6\\
    \{0,27\}  &23.71 &1.37 &0.79 &42.1\\
    \bottomrule
  \end{tabular}
\end{table}
\begin{table}[t]
\renewcommand{\arraystretch}{1.2}
\vspace{5pt}
  \caption{
    Ablation on the impact of scaling synthetic data for underwater dense prediction tasks.
  }
  \label{tab:abl_scaling}
    \centering
    \footnotesize
    \setlength{\tabcolsep}{4mm}
    \begin{tabular}{ccccc}
    \toprule
    $N$ Sample & $SI_{log}\downarrow$ &$A.Rel\downarrow$	&$\delta_{1}\uparrow$ & mIoU\\
    \midrule
    1   & 23.49 &1.46 &0.82 &55.3\\
    3   & 22.94 &1.54 &0.85 &60.9\\
    6   & 22.96 &1.56 &0.85 &63.6\\
    10  & 22.37 &1.50 &0.84 &64.2\\
    \bottomrule
  \end{tabular}
\end{table}

\subsection{Ablation Studies} 
Unless otherwise specified, we conduct ablation studies by training TIDE for 30K iterations. 
We synthesize three samples for each caption, as described in Sec.~\ref{subsec:data synthesis}.
We conduct ablation studies on the USIS10K dataset with SegFormer-B4 for semantic segmentation and the D3 and D5 subsets of the Sea-thru dataset with NewCRFs for depth estimation.\\[5pt]
\noindent \textbf{Ablation on the effectiveness of each component.}
We first evaluate the contribution of each component within TIDE, as shown in Tab.~\ref{tab:abl_components}. 
When utilizing only the Implicit Layout Sharing (ILS) mechanism or Time Adaptive Normalization (TAN), the former outperforms the latter in depth estimation and semantic segmentation.
Combining both methods results in a significant improvement (36.8\% $\rightarrow$ 42.1\%) in semantic segmentation.
These results indicate that ILS and TAN are complementary methods. 
By combining them for end-to-end training, the consistency between images and dense annotations can be further optimized.
Additionally, we further demonstrate the effectiveness of the time-adaptive operation. As shown in the last row of Tab.~\ref{tab:abl_components}, without time-adaptive parameters, the quality of the generated data will be varying degrees of degradation, especially for the semantic segmentation task.\\[5pt]
\noindent \textbf{Ablation on the position of components.}
We then study the effect of the position of ILS and TAN, as shown in Tab.~\ref{tab:abl_position}.
We find that applying the ILS and TAN mechanisms in the first half of the transformer of text-to-image yields better performance than using them in the second half. 
This can be attributed to the layout information produced in the first half of the transformer, which is mismatched with the ILS introduced in the latter part. 
Meanwhile, the results demonstrate that combining both achieves better consistency between the image and dense annotations.\\[5pt]
\noindent \textbf{Ablation on data scaling.}
Finally, we synthesize $N$ samples for each caption to validate the impact of synthetic data scale on underwater dense prediction tasks, as shown in Tab.~\ref{tab:abl_scaling}.
It can be observed that as the amount of synthetic data increases, there is no substantial improvement in the underwater depth estimation task. 
However, for the underwater semantic segmentation task, a significant gain is observed in the early stages as $N$ increases, but the tendency of improvement begins to flatten after $N = 6$.
\subsection{More Challenging Underwater Data Synthesis}
\begin{figure}[t]
  \centering
   \includegraphics[width=1\linewidth]{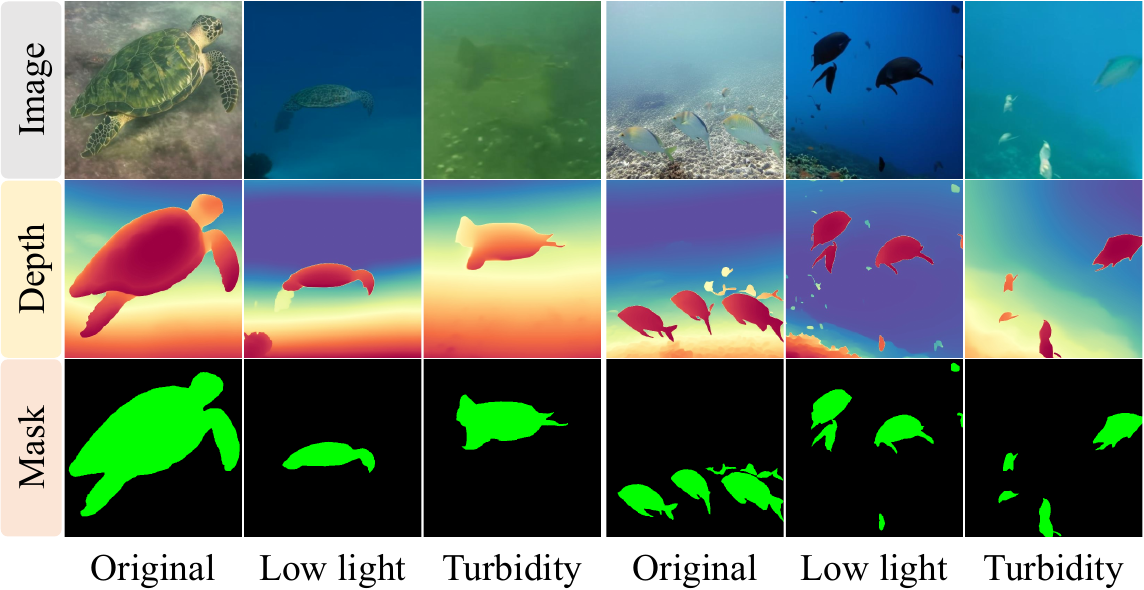}
   \caption{More challenging underwater data generated by TIDE.}
   \label{fig:more_Challenge}
\end{figure}

\begin{figure}[t]
  \centering
   \includegraphics[width=1\linewidth]{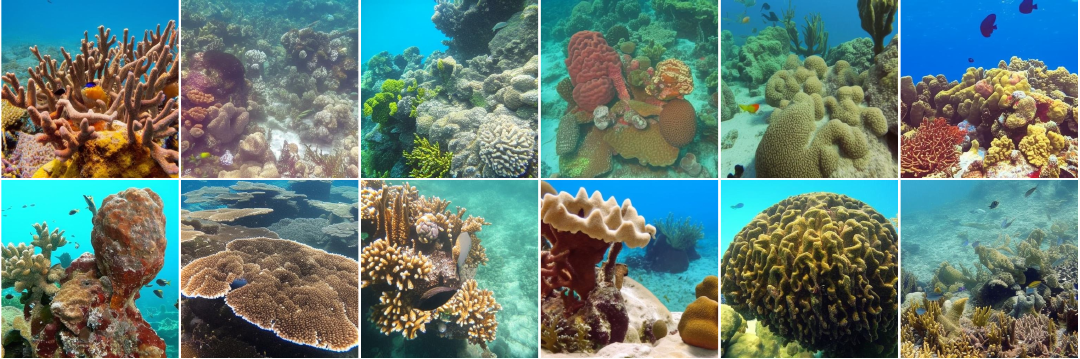}
   \caption{Visualization of generated data diversity.}
   \label{fig:diversity}
      \vspace{-10pt}
\end{figure}

We validate whether TIDE can generate more challenging data by adding extra text prompts about underwater lighting or water quality (e.g., low light, turbidity) to the original underwater scene caption. 
As shown in Fig.~\ref{fig:more_Challenge}, the results demonstrate that TIDE can generate more challenging underwater images. 
While annotating these underwater images may be extremely difficult for humans, TIDE can effortlessly produce highly consistent and accurate dense annotations, which hold great practical value for real-world underwater applications.
In additional, to demonstrate the diversity of generated underwater data, we generate twelve underwater images from the same text prompt, as shown in Fig.~\ref{fig:diversity}. 
It can be observed that, despite sharing the same text prompt, the generated images exhibit rich diversity.
\subsection{Limitation}
Despite the promising results achieved, our method still has some limitations. 
First, our approach cannot generate instance-level semantic masks from the generation perspective. 
Relying on text prompts to guide the generation of instance-level masks with semantic annotations remains challenging. 
Additionally, although TIDE can leverage the powerful priors of pre-trained T2I models to generate highly challenging underwater images (e.g., low light, turbidity), there is still room for improvement. 
These will be key directions for future expansion.
\section{Conclusion}
This paper introduces a unified text-to-image and dense annotation generation model for underwater scenes.
The model can generate realistic underwater images and multiple highly consistent dense annotations using only text prompts as input. 
We validate the effectiveness of our method on underwater depth estimation and semantic segmentation tasks by synthesizing a large-scale underwater dataset.
In the depth estimation task, extensive experiments show that our method, using only text as input, achieves highly competitive results compared to previous methods that required stronger dense conditions for underwater depth synthesis. 
Meanwhile, pre-training with data synthesized using our method further improves model performance in the semantic segmentation task.
Our study provides a new perspective for alleviating data scarcity.

\noindent\textbf{Acknowledgement.} This work was supported by the NSFC (Grant U2341227 and 62225603).

{
    \small
    \bibliographystyle{ieeenat_fullname}
    \bibliography{main}
}

\clearpage
\appendix
\setcounter{figure}{0}
\setcounter{table}{0}
\renewcommand\thefigure{A\arabic{figure}}   
\renewcommand\thetable{A\arabic{table}}
\section{Visualization}

\begin{figure}[h]
  \centering
   \includegraphics[width=1\linewidth]{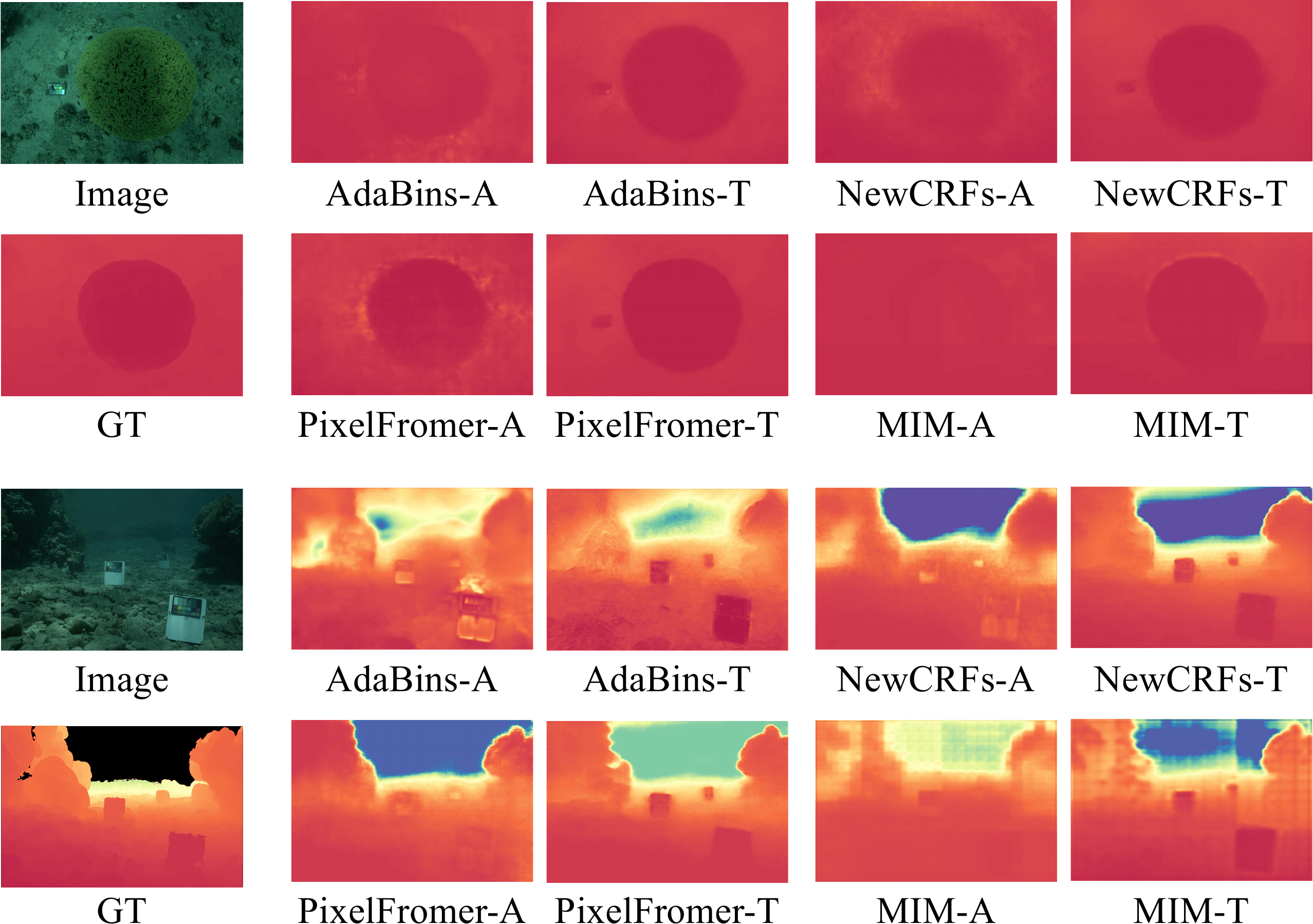}
   \caption{Qualitative results on the Sea-thru dataset~\cite{akkaynak2019sea}. `-A' and `-T' denote models trained on Atlantis~\cite{zhang2024atlantis} and Our SynTIDE dataset, respectively. The depth estimation results are notably improved after training on our dataset. Due to the original `Image' being extremely dim, the content is hardly visible. To clearly display the content of `Image', we adjust its contrast and brightness in this figure. These adjustments do not apply to any inference or evaluation processes at the code level.}
   \label{fig:seathru-vis}
\end{figure}

\begin{figure}[h]
  \centering
   \includegraphics[width=1\linewidth]{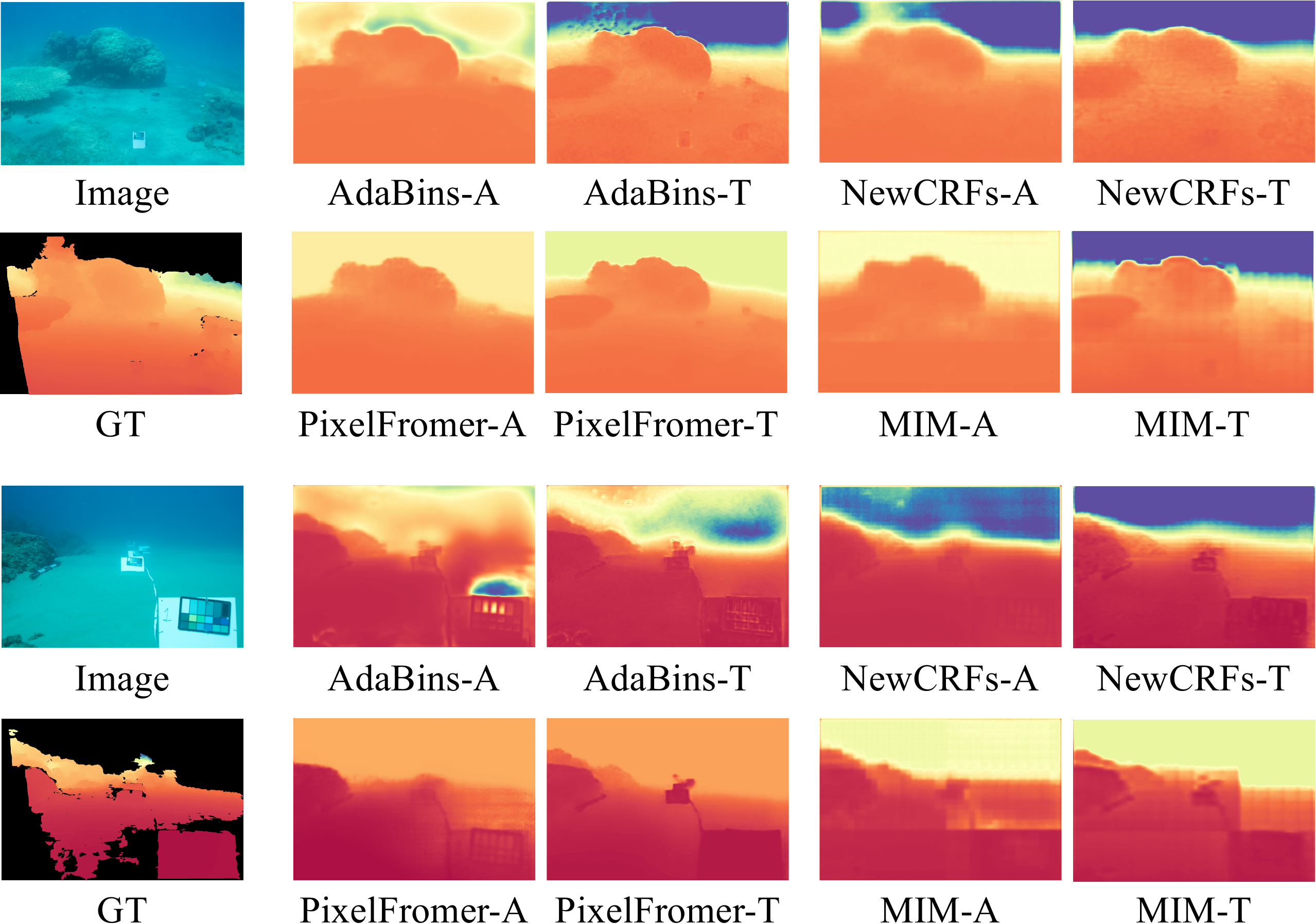}
   \caption{Qualitative results on SQUID dataset~\cite{berman2020underwater}. `-A' and `-T' denote models trained on Atlantis~\cite{zhang2024atlantis} and Our SynTIDE dataset, respectively. The depth estimation results are notably improved after training on our dataset.}
   \label{fig:squid-vis}
\end{figure}

\subsection{Qualitative results}
\label{appendix:Qualitative_results}
Fig.~\ref{fig:seathru-vis} and Fig.~\ref{fig:squid-vis} showcase qualitative comparisons with Atlantis on the D3 and D5 subsets of the Sea-thru~\cite{akkaynak2019sea} dataset and the SQUID~\cite{berman2020underwater} dataset.
All models trained on the SynTIDE dataset, including AdaBins~\cite{bhat2021adabins}, NeWCRFs~\cite{yuan2022neural}, PixerFormer~\cite{agarwal2023attention}, and MIM~\cite{xie2023revealing}, consistently present better visual results on underwater images compared with those trained on the Atlantis dataset. 
Especially in the results of the first two close-shot images in Fig.~\ref{fig:seathru-vis}, the model trained on the Atlantis dataset fails to clearly show the difference in distance between the ball and the background. 
In contrast, our results match the ground truth closer, more distinctly displaying the contrast between the ball and the background in the image.

\subsection{Zero-shot underwater depth data generation}
\label{appendix:Zero-shot}
\begin{figure*}[t]
  \centering
  \includegraphics[width=1\textwidth]{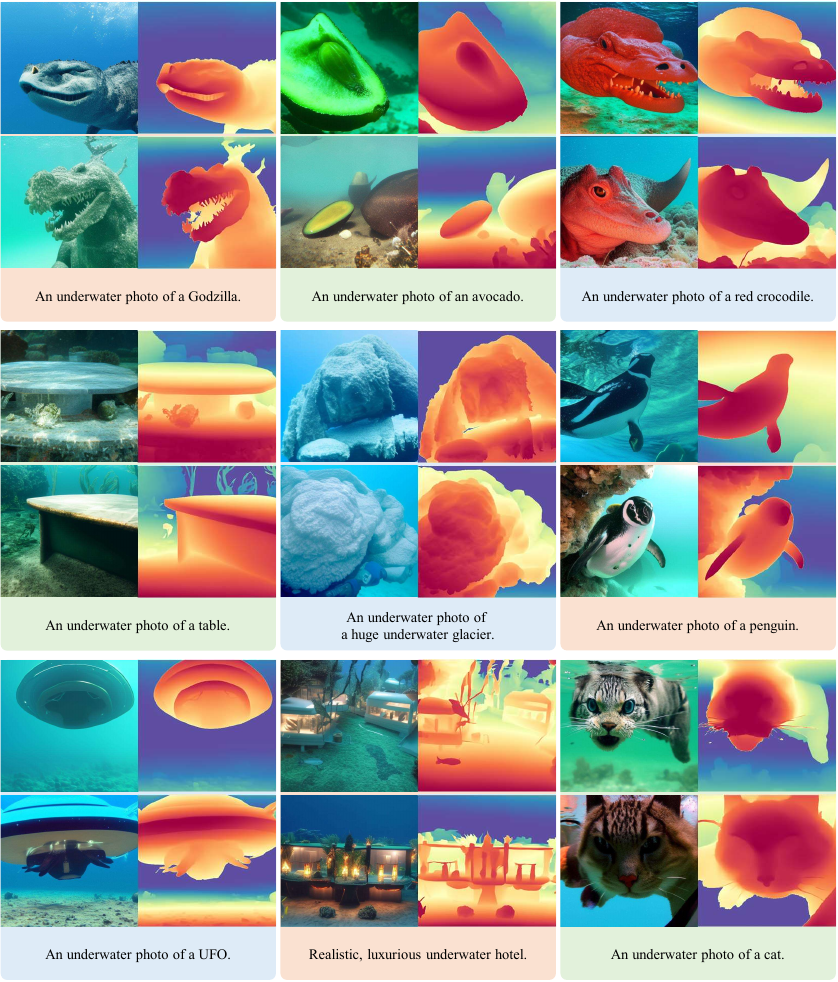}
   \caption{Representative zero-shot image-depth pairs synthesized by TIDE present strong consistency, diversity, and generalization. Images of relevant categories are not included in the training data.}
   \label{fig:zero-vis}
\end{figure*}

Thanks to our training strategy, which fine-tunes the pre-trained text-to-image model~\cite{chen2023pixartalpha} using LoRA~\cite{hu2022lora} with a minor low rank, we retain its strong generalization ability to a certain extent.
This enables TIDE to generate underwater depth data for scenes and objects never seen during training, as shown in Fig.~\ref{fig:zero-vis}.
Even when the provided text prompts contain objects that do not exist in the real world, such as Godzilla, TIDE can still generate seemingly reasonable underwater image-depth pairs. 
However, this capability is particularly challenging for Atlantis~\cite{zhang2024atlantis}, which requires the depth map in advance as a condition.

\end{document}